\documentclass{bmvc2k}

\usepackage{multirow}
\newcommand{\xdownarrow}[1]{%
	{\left\downarrow\vbox to #1{}\right.\kern-\nulldelimiterspace}
}

\usepackage{pifont}
\newcommand{\cmark}{\ding{51}}%
\newcommand{\xmark}{\ding{55}}%

\title{Mobile Vision Transformer-based Visual Object Tracking}

\addauthor{Goutam Yelluru Gopal}{g_yellur@encs.concordia.ca}{1}
\addauthor{Maria A. Amer}{amer@ece.concordia.ca}{1}

\addinstitution{
	Department of Electrical and Computer Engineering\\
	Concordia University, Montr{\'e}al\\
	Qu{\'e}bec, Canada
}

\runninghead{Gopal, Amer}{Mobile Vision Transformer-based Visual Object Tracking}

\def\eg{\emph{e.g}\bmvaOneDot}

\begin{document}
	
\maketitle

\begin{abstract}
The introduction of robust backbones, such as Vision Transformers, has improved the performance of object tracking algorithms in recent years. However, these state-of-the-art trackers are computationally expensive since they have a large number of model parameters and rely on specialized hardware (\eg, GPU) for faster inference. On the other hand, recent lightweight trackers are fast but are less accurate, especially on large-scale datasets. We propose a lightweight, accurate, and fast tracking algorithm using Mobile Vision Transformers (MobileViT) as the backbone for the first time. We also present a novel approach of fusing the template and search region representations in the MobileViT backbone, thereby generating superior feature encoding for target localization. The experimental results show that our MobileViT-based Tracker, \textit{MVT}, surpasses the performance of recent lightweight trackers on the large-scale datasets GOT10k and TrackingNet, and with a high inference speed. In addition, our method outperforms the popular DiMP-50 tracker despite having $4.7\times$ fewer model parameters and running at $2.8\times$ its speed on a GPU. The tracker code and models are available at \url{https://github.com/goutamyg/MVT}. 
\end{abstract}

\section{Introduction}\label{sec:intro}
The two prominent paradigms of visual object tracking algorithms are Discriminative Correlation Filters (DCFs) and deep Siamese Networks (SNs) \cite{9913708}. The DCF-based trackers \cite{henriques2014high, danelljan2017eco, bhat2019learning} localize the target object based on the filter response generated by convolving the features extracted from the search region with the filter coefficients learned from the target template. The SN-based trackers \cite{yan2021learning, chen2021transformer, lin2022swintrack} perform the cross-correlation (or similar) operation between features extracted from the template and search regions to generate the response map for target localization and bounding box estimation. The explicit learning of target-specific filter coefficients in DCF tracking increases their robustness against semantic background regions compared to SN trackers; however, SNs are faster due to their simpler model architecture supporting end-to-end evaluation on a GPU. With the adoption of powerful backbones and effective feature fusion techniques, SN trackers have shown state-of-the-art performance on various benchmarks \cite{fan2021lasot, muller2018trackingnet, Huang2021}.

Feature representation of the target object plays a crucial role in tracker performance \cite{wang2015understanding}. Most SN trackers use the ResNet \cite{he2016deep} backbone for feature extraction, with ResNet-50 and ResNet-101 being the popular choices. The more recent trackers use pre-trained Vision Transformer (ViT) models \cite{dosovitskiy2021an, wu2021cvt} as their backbone, surpassing the performance of ResNet-based SN trackers. However, a notable disadvantage of ViT-based trackers is the complexity of their backbone, both in terms of memory (a large number of model parameters) and latency (low inference speed). By deploying these models, achieving high tracking speed on an ordinary CPU or mobile device is challenging. This limitation severely restricts the usage of such tracking algorithms for several resource-constrained applications. 

On the other hand, most lightweight tracking algorithms deploy compact convolutional neural network (CNN) backbones to minimize the model latency. The inductive biases of convolutional blocks effectively model the spatially local information related to the target object but fail to capture the global relations essential for accurate target state estimation in tracking \cite{yan2021learning}. Such a lack of global association between the template and search region in the backbone increases the burden on the feature fusion module (or the neck) to generate the fused encoding favorable for accurate and robust tracking. The self-attention-based Transformers \cite{vaswani2017attention} as the backbone is effective at global contextual modeling and have been excellent for tracking \cite{cui2022mixformer, ye2022joint}; however, they are computationally expensive. 

In this paper, we are the first to investigate the usefulness of Mobile Vision Transformers (MobileViTs) as the backbone for single object tracking to present a lightweight but high-performance tracking algorithm, \textit{MVT}. The recent MobileViTs \cite{mehta2022mobilevit} for image classification are known for their low latency, lightweight architecture, and adaptability to downstream tasks, \eg, object detection and semantic segmentation. In addition, while all the related lightweight trackers independently compute the template and search region features in their respective backbone, our \textit{MVT} algorithm employs a hybrid feature extraction method where template and search regions are blended in the backbone by our novel Siamese Mobile Vision Transformer (Siam-MoViT) block.

\section{Related Work}\label{sec:related_work}
Multiple SN-based lightweight trackers have been presented in the last few years. LightTrack \cite{yan2021lighttrack} employed Neural Architecture Search \cite{chen2019detnas} to present an efficient tracking pipeline. It designed a search space of lightweight building blocks to find the optimal backbone and head architectures with pre-set constraints on the number of model parameters. E.T.Track \cite{blatter2023efficient} incorporated Exemplar Transformers for tracking to achieve real-time speed on a CPU. It used a stack of lightweight transformer blocks in the head module to perform target classification and bounding box regression. FEAR \cite{borsuk2022fear} tracker deployed a dual-template representation to incorporate temporal information during tracking. With a compact backbone, FEAR achieved over 200 frames-per-second (\textit{fps}) speed on iPhone 11 with negligible impact on battery level. Stark-Lightning \cite{yan2021learning} used a RepVGG \cite{ding2021repvgg} backbone and a transformer-based encoder-decoder architecture in the neck module to model spatio-temporal feature dependencies between the target template and search regions. HiFT \cite{cao2021hift} proposed a hierarchical feature transformer-based approach for aerial tracking. It generated hierarchical similarity maps from the multi-level convolutional layers in the backbone network to perform a transformer-based fusion of shallow and deep features. SiamHFFT \cite{dai2022siamese} extended the hierarchical feature fusion approach by \cite{cao2021hift} to model the inter-dependencies within the multi-level features and achieve high tracking speed on a CPU.

Among the related lightweight trackers, LightTrack is closest to our work, having similar neck and head modules but a different backbone. Stark-Lightning uses a transformer-based neck module to fuse features from the template and search regions. In contrast, the proposed \textit{MVT} uses a simple, parameter-free cross-correlation operation in its neck module. E.T.Track uses a transformer-based head module, while our \textit{MVT}'s head module is built using a fully convolutional network. As a post-processing step, the related trackers LightTrack and E.T.Track refine their predicted bounding boxes by penalizing significant changes in bounding box size and aspect ratio between consecutive frames. Unlike these trackers, the proposed \textit{MVT} does not perform such heuristic-based bounding box refinements. 

\begin{figure}
	\includegraphics[width=\textwidth]{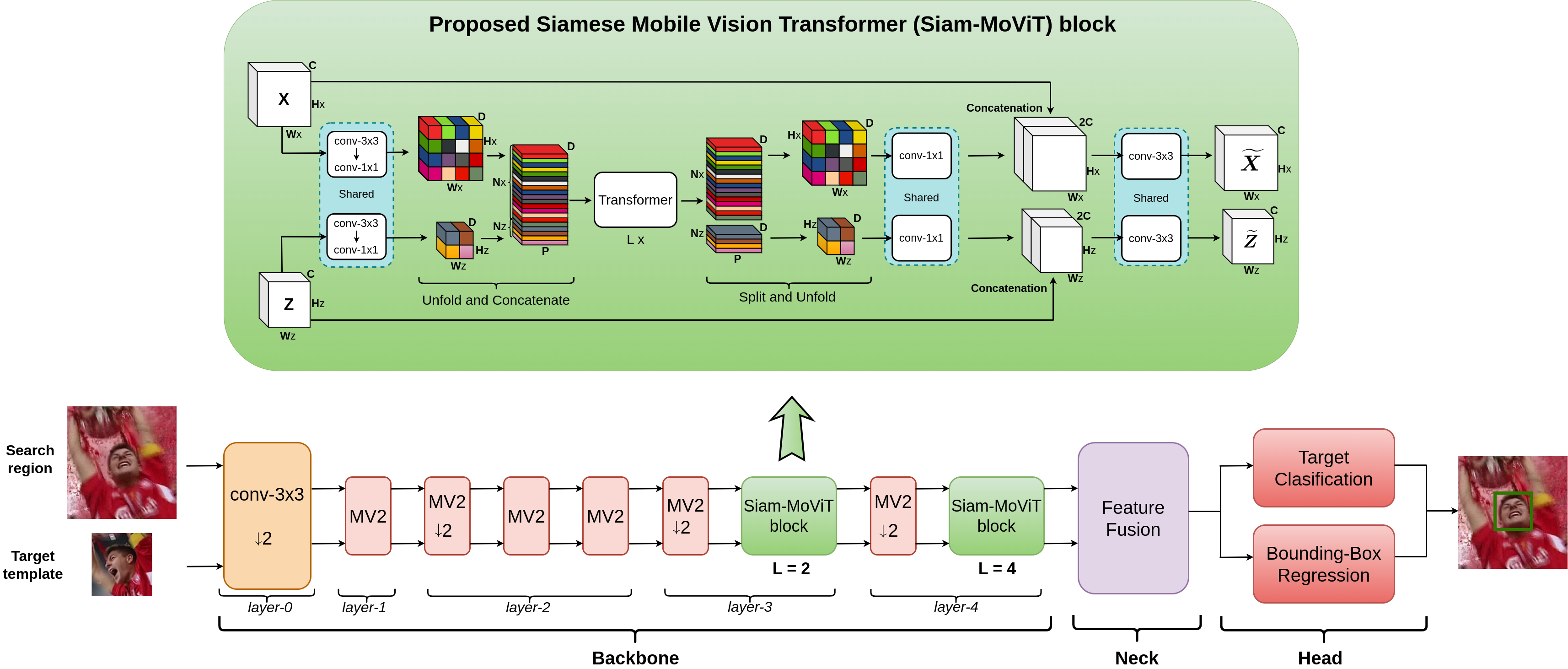}
	\caption{The pipeline of the proposed \textit{MVT} tracker and our \colorbox{green!30}{Siam-MoViT} block. The backbone consists of MobileNetV2 \cite{sandler2018mobilenetv2} (or \colorbox{red!20}{MV2}) and \colorbox{green!30}{Siam-MoViT} blocks for feature extraction. $\xdownarrow{1.5mm}$2 indicates spatial downsampling by a factor of 2. Details of our Siam-MoViT block can be found in Section \ref{sec:proposed_method}.}
	\label{fig:MVT-architecture}
\end{figure}

Most importantly, all the related lightweight trackers use a two-stream approach during feature extraction, i.e., the backbone features from the template and search region are computed independently. Such a two-stream computation limits the interaction between the template and search regions to the neck module only, resulting in inferior tracking performance. To alleviate this problem, we propose a hybrid feature extraction method where template and search regions are blended in the backbone by our novel Siam-MoViT block, as shown in Figure \ref{fig:MVT-architecture}. The resulting entangled feature representation generated using our Siam-MoViT block improves the tracker performance while maintaining high inference speed. Efficient transformer architectures is an emerging research topic \cite{tay2022efficient} and has been unexplored by previously proposed lightweight trackers. To our knowledge, we are the first to use MobileViT as the backbone for object tracking. We are also the first to propose a tracking pipeline with a joint feature extraction and fusion approach in the tracker backbone. 

Our contributions in this paper are thus:
\begin{itemize}
	\item A novel lightweight tracking algorithm using MobileViTs. We show that the proposed MobileViT-based tracker performs better than related lightweight trackers. 
	\item A hybrid feature extraction approach, intertwining the template and search regions using our Siam-MoViT block, producing better features for target state estimation.
\end{itemize}

\section{Proposed Mobile Vision Transformer-based Tracker}\label{sec:proposed_method}
In this section, we discuss the pipeline of our \textit{MVT} algorithm for single object tracking (shown in Figure \ref{fig:MVT-architecture}) and information related to model training. 

\subsection{Proposed \textit{MVT} Backbone and the Siam-MoViT block}\label{sec:backbone}
The input to our \textit{MVT} backbone is a pair of the target template and search region image patches, $Z_{in} \in R^{W_z \times H_z \times 3}$ and $X_{in} \in R^{W_x \times H_x \times 3}$, respectively. The tracker backbone consists of cascaded MobileNetV2 \cite{sandler2018mobilenetv2} and the proposed Siam-MoViT blocks, as shown in Figure \ref{fig:MVT-architecture}. These modules process the input image patches sequentially, with recurrent spatial down-sampling operations to reduce the feature dimensionality. The proposed Siame-MoViT block uses a modified MobileViT block \cite{mehta2022mobilevit}, especially around the transformer encoder, to accommodate features from the template and search region. 

Our Siam-MoViT block receives a pair of intermediate feature maps $Z$ and $X$, belonging to the template and search regions, respectively. We assume that $Z$ and $X$ have $C$ channels. Inside the Siam-MoViT block, first, we apply a $3\times3$ convolutional filter to learn spatially local feature representations. It is followed by a $1\times1$ convolutional filter, projecting the features onto a $D$-dimensional space as a linear combination of $C$ input channels. Next, we perform the \textit{unfold and concatenate} operation (\textit{cf.} Figure \ref{fig:MVT-architecture}), where we divide the feature maps $X$ and $Z$ into $N$ non-overlapping patches of size $w \times h$. We then flatten these patches to generate tokens of size $P \times N \times D$, where $P = w \cdot h$ and $N = \frac{W \cdot H}{P}$. These tokens are concatenated and passed through a series of $L$ transformer blocks to encode the global relationship between the template and search regions. Our implementation uses the standard multi-headed self-attention transformer encoder blocks \cite{vaswani2017attention}. This operation of learning self-attention on the concatenated features facilitates the exchange of information between template and search regions, thereby generating high-quality encodings for robust target localization. To restore the spatial ordering of feature maps, we split the output tokens from the transformer and re-arrange them to obtain feature maps of size $H_z \times W_z \times D$ and $H_x \times W_x \times D$, shown as the \textit{split and unfold} operation in Figure \ref{fig:MVT-architecture}. Then, we re-map the number of channels from $D$ to $C$ by applying a $1 \times 1$ convolutional filter and concatenate the resulting feature maps with the inputs to the Siam-MoViT block, i.e., $Z$ and $X$. Finally, we apply a $3 \times 3$ convolutional filter on the concatenated feature maps to generate the output of our Siam-MoViT block, denoted as $\tilde{Z}$ and $\tilde{X}$, having the same size as $Z$ and $X$, respectively. Note that all the MobileNetV2 blocks in the backbone and the CNN blocks within the Siam-MoViT block are applied separately to template and search regions, as shown in Figure \ref{fig:MVT-architecture}, with shared weights. 

\subsection{Neck and Head Modules}\label{sec:neck_and_head}
The output from the last layer of the \textit{MVT} backbone has feature maps corresponding to the template and search region. We fuse these features in the neck module to generate an encoded feature representation $f_{zx} \in R^{\frac{H_z \cdot W_z}{16^2} \times \frac{H_x}{16} \times \frac{H_x}{16}}$ for target state estimation. For this, we use a simple pointwise cross-correlation operator \cite{yan2021alpha} in the neck module, the same as LightTrack \cite{yan2021lighttrack}. We use a layer of batch-normalization (BN) \cite{ioffe2015batch} before performing cross-correlation. We then apply a $1 \times 1$ convolutional \textit{channel-adjust} layer on $f_{zx}$ to match the number of channels between $f_{zx}$ and the head module. 

For classification and regression, we adopt the head module from \cite{ye2022joint}, which uses a fully convolutional network (FCN) to perform target classification and bounding box regression. The FCN consists of a stack of five Conv-BN-ReLU blocks. The classification network predicts a score map $\mathcal{R} \in R^{\frac{H_x}{16} \times \frac{W_x}{16}}$, and the position of the maximum value in $\mathcal{R}$ is considered as the target location. The regressor network predicts the normalized bounding box size (i.e., target width and height) and corresponding local offset values.

\subsection{Loss Function for Training}
During training, we use loss functions for the classification and regression output by the head module of our \textit{MVT} tracker. As in \cite{ye2022joint}, we use the weighted focal loss $L_{cls}$ to handle the imbalance between positive and negative training examples for target classification. For bounding box regression, same as \cite{ye2022joint}, we use the $\ell_1$ and generalized IoU loss functions, denoted by $L_1$ and $L_{giou}$, respectively. As in \cite{ye2022joint}, we define the overall loss function as,
\begin{equation}\label{eq:loss_function}
	L_{total} = L_{cls} + \lambda_1 \cdot L_1 + \lambda_2 \cdot L_{giou},
\end{equation}
where $\lambda_1$ and $\lambda_2$ are the hyperparameters controlling the relative impact of $L_1$ and $L_{giou}$ on the overall training loss. 


\section{Implementation Details and Experimental Results}
This section discusses the implementation details of our \textit{MVT} tracker and compares its results with related lightweight and state-of-the-art heavy trackers. We also discuss the ablation study results for the proposed feature fusion and an attribute-based robustness analysis.

\subsection{Implementation Details}
We set the dimensions of the inputs to our \textit{MVT} backbone, i.e., $Z_{in}$ and $X_{in}$ from Section \ref{sec:backbone}, to $128 \times 128$ and $256 \times 256$, respectively. We divide our \textit{MVT} backbone into five layers with $layer$-$id$s for notation convenience, as shown in Figure \ref{fig:MVT-architecture}. The number of channels in the feature maps increases along these five layers as $\{3 \to 16, 16 \to 32, 32 \to 64, 64 \to 96, 96 \to 128\}$. We set the number of transformer encoders for the proposed Siam-MoViT block in $layer$-$3$ and $layer$-$4$ to 2 and 4, respectively. We set the parameters $w = h = 2$ for folding and unfolding operations within our Siam-MoViT block. The number of upscaled channels $D$ in the Siam-MoViT block is set to 144 and 192 for $layer$-$3$ and $layer$-$4$, respectively. The backbone has a stride of 16 (i.e., four downsampling operations, each by a factor of two), resulting in feature maps of size $8 \times 8$ and $16 \times 16$ for the template and search regions, respectively. The \textit{channel-adjust} layer in the neck module, described in Section \ref{sec:neck_and_head}, upscales the number of channels from 64 to 256. 

We use the training split of the GOT10k dataset \cite{Huang2021} to train our model. We use \textit{Adam-W} \cite{loshchilov2018decoupled} as the optimizer with a weight decay of $10^{-4}$. We trained our model for 100 epochs with 60000 image pairs per epoch, sampled from the training dataset. We use the validation split of GOT10k to compute the values of $L_{cls}$, $L_1$, and $L_{giou}$ from Eq. \ref{eq:loss_function} during training to examine the possibility of overfitting. We set the initial learning rate \textit{lr} to $4 \times 10^{-4}$ and use cosine annealing \cite{loshchilov2017sgdr} as the learning rate scheduler (without the warm restarts). We keep the \textit{lr} for the backbone module 0.1 times the \textit{lr} for the rest of the network throughout training. We use the data augmentation techniques horizontal flip and brightness jitter during training. We initialize the backbone using the weights of the pre-trained MobileViT model provided by its authors \cite{mehta2022mobilevit}. Like \cite{mehta2022mobilevit}, we do not use positional embeddings for the transformer blocks in our \textit{MVT} backbone. We set the hyperparameters $\lambda_1$ and $\lambda_2$ in Eq. \ref{eq:loss_function} to 5 and 2, respectively, as in \cite{ye2022joint}. We use a single Nvidia Telsa V100 GPU (32GB) for training and set the batch size to 128. 

Our choice of optimizer and hyperparameters is based on the training settings typically used by the related trackers. We set our batch size based on the maximum number of images that can be loaded onto the GPU used for training the model. We experimented using the Ray-Tuner package in Pytorch \cite{paszke2019pytorch} to search for the best set of hyperparameters jointly. First, the hyperparameter search was time-consuming due to the sheer volume. Second, due to a strong inter-dependency between some of the hyperparameters (\eg, batch size and learning rate), it was challenging to find the optimal set using random search-based methods.

During inference, we define the search space at frame $t$ by extracting an image patch around the estimated target location at frame $t-1$, four times the area of the target template. We apply a Hanning window on the classification score map $\mathcal{R}$ as the post-processing step. After this multiplication, we determine the location of the highest value in $\mathcal{R}$ as the target location, and we choose the corresponding bounding box as the tracker output. We define the target annotation from the first frame as the template and do not perform any model update. We generate the GPU-based inference results using an Nvidia RTX 3090 GPU.  


\subsection{Results and Comparison to Related Work}\label{subsec:comparison_to_related_work}
To demonstrate the effectiveness of the proposed \textit{MVT}, we evaluate it using GOT10k-test \cite{Huang2021}, TrackingNet-test \cite{muller2018trackingnet}, LaSOT-test \cite{fan2021lasot}, and NfS30 \cite{kiani2017need} datasets. GOT10k has 180 test videos, with non-overlapping target classes from their training videos, to promote generalization during tracker development. TrackingNet has 511  challenging test videos with 15 attributes. GOT10k and TrackingNet datasets sequester the test set annotations and provide an online evaluation server to submit the tracker results to ensure a fair evaluation. LaSOT dataset has 280 test videos, with an average length of 2500 frames per video. NfS dataset has 100 videos captured at 240 and 30 \textit{fps}; we use the 30 \textit{fps} videos. GOT10k provides a train, validation, and test split for its annotated videos, whereas TrackingNet and LaSOT provide the train and test splits. NfS30 has only test videos in the dataset.

GOT10k uses Overlap Ratio ($OR$) and Success Rate ($SR$) at a threshold of 0.5 and 0.75 (i.e., $SR_{0.50}$ and $SR_{0.75}$) to quantify the tracker performance. Metric $OR$ is equivalent to Area Under the Curve ($AUC$) \cite{vcehovin2016visual}. $SR$ measures the fraction of frames where the Intersection-over-Union ($IoU$) between groundtruth and predicted boxes is higher than a threshold. TrackingNet uses $AUC$, Precision ($P$), and Normalized-Precison ($P_{norm}$) for tracker performance. Precision $P$ measures the distance between centers of groundtruth and predicted bounding boxes, whereas $P_{norm}$ computes the same metric using normalized bounding boxes. For LaSOT and NfS30, we use the $AUC$ and Failure Rate ($FR$) as the performance metrics. $FR$ calculates the fraction of frames where the tracker has drifted away, i.e., its bounding box prediction has no overlap with the groundtruth (i.e., $IoU$ score is zero). 

We compare the results of the proposed \textit{MVT} with the related lightweight trackers: LightTrack \cite{yan2021lighttrack}, Stark-Lightning \cite{yan2021learning}, FEAR-XS \cite{borsuk2022fear}, and E.T.Track \cite{blatter2023efficient}, evaluated using the pretrained models provided by their authors. From Table \ref{table:qualitative_results}, we can see that our \textit{MVT} outperforms all other lightweight trackers on the server-based test set of GOT10k and TrackingNet. 
\begin{table}[t]
	\centering
	\resizebox{\textwidth}{!}{
		\begin{tabular}{c|ccc|ccc||cc|cc||c}
			\hline
			\multicolumn{1}{c|}{Tracker} & \multicolumn{3}{c|}{\textbf{GOT10k} \cite{Huang2021} (server)} & \multicolumn{3}{c||}{\textbf{TrackingNet} \cite{muller2018trackingnet} (server)} & \multicolumn{2}{c|}{NfS30 \cite{kiani2017need}} & \multicolumn{2}{c||}{LaSOT \cite{fan2021lasot}} & \multicolumn{1}{c}{\textit{fps}}\\
			\multicolumn{1}{c|}{} & $OR \uparrow$ & $SR_{0.50} \uparrow$ & $SR_{0.75} \uparrow$ & $AUC \uparrow$ & $P_{norm} \uparrow$ & $P \uparrow$ & $AUC \uparrow$ & $FR \downarrow$ & $AUC \uparrow$ & $FR \downarrow$ & (GPU)\\
			\hline
			LightTrack \cite{yan2021lighttrack} (CVPR'21) & 0.582 & 0.668 & 0.442 & 72.9 & 79.3 & {\color{blue}69.9} & 0.582 & 0.146 & 0.524 & {\color{red}\textbf{0.116}} & 99 \\
			Stark-Lightning \cite{yan2021learning} (ICCV'21) & {\color{blue}0.596} & {\color{blue}0.696} & {\color{blue}0.479} & 72.7 & 77.9 & 67.4 & {\color{red}\textbf{0.619}} & {\color{blue}0.111} & {\color{blue}0.585} & 0.151 & 205 \\
			FEAR-XS \cite{borsuk2022fear} (ECCV'22) & 0.573 & 0.681 & 0.455 & 71.5 & {\color{blue}80.5} & {\color{blue}69.9} & 0.487 & 0.207 & 0.508 & 0.273 & 275 \\
			E.T.Track \cite{blatter2023efficient} (WACV'23) & 0.566 & 0.646 & 0.425 & {\color{blue}74.0} & 79.8 & 69.8 & 0.589 & 0.172 & {\color{red}\textbf{0.597}} & 0.162 & 53 \\
			MVT (ours) & {\color{red}\textbf{0.633}} & {\color{red}\textbf{0.742}} & {\color{red}\textbf{0.551}} & {\color{red}\textbf{74.8}} & {\color{red}\textbf{81.5}} & {\color{red}\textbf{71.9}} & {\color{blue}0.603} & {\color{red}\textbf{0.085}} & 0.553 & {\color{blue}0.137} & 175 \\
			\hline
	\end{tabular}} 
	\caption{Comparison of related lightweight SN trackers with our \textit{MVT} on server-based GOT10k-test and TrackingNet-test, and groundtruth available NfS30 and LaSOT-test datasets. The best and second-best results are highlighted in {\color{red}red} and {\color{blue}blue}, respectively.}
	\label{table:qualitative_results}
\end{table}
No related tracker scores second best constantly for these datasets. On GOT10k-test, our tracker is better by at least 3.7\%, 4.6\%, and 7.3\%, than the second best tracker in terms of $OR$, $SR_{0.50}$, and $SR_{0.75}$, respectively. Recall that GOT10k-test has unseen object classes; this indicates a higher generalization ability of \textit{MVT} towards tracking novel object classes than the related trackers. It also highlights the impact of feature fusion in our tracker backbone compared to other two-stream-based lightweight trackers. We observe a similar behavior using the TrackingNet dataset, where our \textit{MVT} performs better by approximately $2\%$ in AUC, $P$, and $P_{norm}$ than its competitor, LightTrack. No single tracker constantly performs better in $AUC$ or $FR$ for the NfS30 and LaSOT datasets with groundtruth available for the test sets. For NfS30, our tracker is better by 2.6\% in $FR$ than the second-best Stark-Lightning while lower by 1.6\% in AUC. For LaSOT, our tracker is lower by 2.1\% than the second best LightTrack in $FR$ and by 4.4\% than the best E.T.Track in $AUC$. 

Across all the datasets and performance metrics, we can see that our tracker scores the best in most cases (7/10) while being second best in 2/10 cases. Our closest competitor, Stark-Lightning, scores the second-best 5/10 times and the best only once. Regarding speed, our \textit{MVT} runs 175 \textit{fps} during GPU-based evaluation, that is, 15\% slower than its competitor Stark-Lightning, as shown in Table \ref{table:qualitative_results}. It is because Stark-Lightning computes the template region features only once during inference due to its two-stream tracking pipeline. In contrast, our \textit{MVT} requires evaluation of the template features at every frame due to the entanglement of the template and search regions in its backbone, which impacts tracking speed. 

\subsection{Comparison to State-of-the-art trackers}
In Table \ref{table:sota_comparison}, we compare the proposed \textit{MVT} to state-of-the-art (SOTA) heavyweight trackers on server-based GOT10k and TrackingNet test datasets. We take the values of evaluation metrics for these trackers from the respective papers; however, we compute their \textit{fps} values on a GPU (i.e., Nvidia RTX 3090) and a CPU (i.e., 12th Gen Intel(R) Core-i9 processor), as shown in the last column of Table \ref{table:sota_comparison}. As we can see, in comparison to the popular DCF-based DiMP-50 \cite{bhat2019learning}, the deployment of transformers for feature fusion \cite{chen2021transformer, yan2021learning} and as the backbone \cite{ye2022joint, cui2022mixformer} has improved the tracker performance, but at the cost of increased computational complexity and lowered tracking speed due to higher number of model parameters. In contrast, proposed \textit{MVT} surpasses the performance of the popular DiMP-50 on GOT10k and TrackingNet datasets with $4.7 \times$ fewer parameters while running at $2.8 \times$ and $2 \times$ its speed on a GPU and CPU, respectively. Compared to the best-performing SOTA tracker MixFormer-L \cite{cui2022mixformer} in Table \ref{table:sota_comparison}, proposed \textit{MVT} has $33.43 \times$ fewer model parameters and higher \textit{fps}, i.e., $3.87 \times$ on GPU and $5.88 \times$ on CPU, but has a lower $AUC$ of 10.7\% on average across the two datasets. Our tracker provides a tradeoff between accuracy and complexity for real-time applications with resource constraints.
\begin{table}[tbh]
	\centering
	\resizebox{\columnwidth}{!}{
		\begin{tabular}{c|cc|cc||c|cc}
			\hline
			\multicolumn{1}{c|}{Tracker} & \multicolumn{2}{c|}{GOT10k} & \multicolumn{2}{c||}{TrackingNet} & \multicolumn{1}{c|}{$\#$params $\downarrow$} & \multicolumn{2}{c}{\textit{fps}}\\
			\multicolumn{1}{c|}{} & $OR$ $\uparrow$ & $SR_{0.50} \uparrow $ & $AUC$ $\uparrow$& $P_{norm}$ $\uparrow$ & (in millions) & GPU $\uparrow$ & CPU $\uparrow$\\
			\hline
			DiMP-50 \cite{bhat2019learning} & 0.611 & 0.717 & 74.0 & 80.1 & 26.1 & 61.5 & {\color{blue}15.0} \\
			
			TransT \cite{chen2021transformer} & 0.671 & 0.768 & 81.2 & 85.4 & {\color{blue}23.0} & {\color{blue}87.7} & 2.3\\
			
			STARK-ST101 \cite{yan2021learning} & 0.688 & 0.781 & {\color{blue}82.0} & 86.9 & 47.2 & 80 & 7.8\\
			
			OSTrack-384 \cite{ye2022joint} & {\color{blue}0.740} & {\color{blue}0.835} & \textbf{\color{red}83.9} & {\color{blue}88.5} & 92.1 & 74.4 &  4.4\\
			
			MixFormer-L \cite{cui2022mixformer} & \textbf{\color{red}0.756} & \textbf{\color{red}0.857} & \textbf{\color{red}83.9} & \textbf{\color{red}88.9} & 183.9 & 45.2 & $<$ 5\\
			
			\hline
			\hline
			MVT (ours) & 0.633 & 0.742 & 74.8 & 81.5 & \textbf{\color{red}5.5} & \textbf{\color{red}175.0} & \textbf{\color{red}29.4}\\
			\hline
	\end{tabular}} 
	\caption{Comparison of our \textit{MVT} with the state-of-the-art heavyweight trackers on server-based GOT10k and TrackingNet test datasets. Best and second best results in accuracy and complexity (i.e., $\#$ of parameters and \textit{fps}) are highlighted in {\color{red}red} and {\color{blue}blue}, respectively.}
	\label{table:sota_comparison}
\end{table}
%

\subsection{Ablation Study}
To analyze the effectiveness of the proposed feature fusion technique deployed in our \textit{MVT} backbone, we evaluate the performance of our tracker trained without the concatenation of the template and search region features inside the proposed Siam-MoViT block (\textit{cf.} Figure \ref{fig:MVT-architecture}). Table \ref{table:ablation_results} summarizes the ablation results on the four datasets discussed in Section \ref{subsec:comparison_to_related_work}. We can see that the proposed feature fusion improves the $OR$ (or the equivalent metric $AUC$) by 1.9\% on average across all the datasets. It also increases the robustness of our \textit{MVT} tracker by reducing the $FR$ on NfS30 and LaSOT datasets by 3.7\% and 2.6\%, respectively. Learning self-attention on the concatenated features using the transformer blocks in our \textit{MVT} backbone facilitates the global relational modeling \textit{within} and \textit{between} the template and search regions, thereby generating superior features for accurate target localization and robust tracking.
\begin{table}[t]
	\centering
	\begin{tabular}{c|cc|cc|cc|cc}
		\hline
		\multicolumn{1}{c|}{feature fusion} & \multicolumn{2}{c|}{GOT10k} & \multicolumn{2}{c|}{TrackingNet} & \multicolumn{2}{c|}{NfS30} & \multicolumn{2}{c}{LaSOT} \\
		\multicolumn{1}{c|}{in backbone} & $OR$ $\uparrow$ & $SR_{0.50} \uparrow $ & $AUC$ $\uparrow$& $P_{norm}$ $\uparrow$ & $AUC$ $\uparrow$& $FR$ $\downarrow$& $AUC$ $\uparrow$& $FR \downarrow$\\
		\hline
		\xmark  \hspace{8mm} & 0.600 & 0.703 & {\color{red}\textbf{74.9}} & 80.0 & 0.566 & 0.122 & 0.544 & 0.163 \\
		\cmark (ours) & {\color{red}\textbf{0.633}} & {\color{red}\textbf{0.742}} & 74.8 & {\color{red}\textbf{81.5}} & {\color{red}\textbf{0.603}} & {\color{red}\textbf{0.085}} & {\color{red}\textbf{0.553}} & {\color{red}\textbf{0.137}} \\
		\hline
	\end{tabular}
	\caption{Ablation study results related to the proposed feature fusion in our \textit{MVT} backbone. Best results are highlighted in {\color{red}red}.}
	\label{table:ablation_results}
\end{table}

\subsection{Robustness Analysis}
To analyze the robustness of the proposed \textit{MVT} tracker against various challenging factors (or attributes), we compute its $FR$ for attributes annotated under the LaSOT dataset, namely Aspect Ration Change (\textit{ARC}), Background Clutter (\textit{BC}), Camera Motion (\textit{CM}), Deformation (\textit{DEF}), Fast Motion (\textit{FM}), Full Occlusion (\textit{FOC}), Illumination Variation (\textit{IV}), Low Resolution (\textit{LR}), Motion Blur (\textit{MB}), Out-of-View (\textit{OV}), Partial Occlusion (\textit{POC}), Rotation (\textit{ROT}), Scale Variation (\textit{SV}), and Viewpoint Change (\textit{VC}). 
\begin{figure}[tbh]
	\centering
	\includegraphics[width=0.8\textwidth]{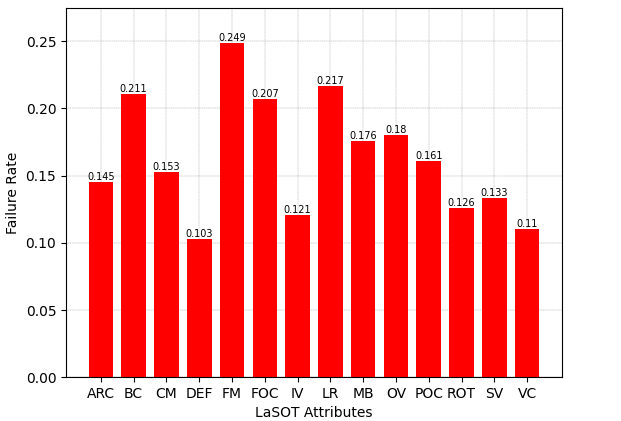}
	\caption{Analyzing the robustness of our \textit{MVT} based on its failure rate $FR$ for different attributes on the LaSOT test dataset. The average $FR$ is 0.137.}
	\label{fig:robustness_analysis}
\end{figure}
From Figure \ref{fig:robustness_analysis}, we can see that our \textit{MVT} is most robust to target deformation (\textit{DEF}) and appearance changes (\textit{VC}). It is least robust to attribute \textit{FM} since we use a Hanning window on the classification score map during target localization. However, not using the Hanning window deteriorates the robustness of our tracker against \textit{BC} and increases the overall \textit{FR}, as we observed from our experiments. Also, our \textit{MVT} has a higher $FR$ for videos under the attribute \textit{LR}. These videos contain small, texture-less target objects such as \textit{volleyball} and \textit{yo-yo}, which are generally fast-moving (i.e., \textit{FM}) and are sensitive to \textit{BC}. SOTA trackers address the challenges of \textit{FM}, \textit{LR}, and \textit{BC} with deep features and larger search area to avoid target loss, but these improvements come at the expense of higher model complexity and memory footprint, as shown in Table \ref{table:sota_comparison}. 

\section{Conclusion and Future Work}
In this paper, we proposed \textit{MVT}, our visual object tracking algorithm that uses, for the first time, the Mobile Vision Transformers as the backbone. We also proposed the Siam-MoViT block to model the global interactions between template and search regions in the tracker backbone, thereby enhancing the quality of feature encodings for target localization. Our simulation results showed that the proposed tracker performed better than the related lightweight trackers on the large-scale GOT10k and TrackingNet datasets, showcasing the effectiveness of the proposed tracking method. Despite having $4.7 \times$ fewer model parameters, our \textit{MVT} performs better than the popular DCF-based DiMP-50 tracker, while running at least $2 \times$ its speed during CPU and GPU-based evaluation. Our ablation studies highlighted the importance of the proposed feature fusion on our tracker performance.

In our future work, we plan to explore lightweight vision transformer backbone architectures to enhance the quality of encoded features further. Effective feature fusion in the backbone can make the neck module redundant for lightweight tracking, simplifying the tracking pipeline. We also plan to deploy and test our models on low-memory embedded devices, such as smartphones. 

\bibliography{strings,Goutamrefs_abbrv}

\begin{thebibliography}{34}
\providecommand{\natexlab}[1]{#1}
\providecommand{\url}[1]{\texttt{#1}}
\expandafter\ifx\csname urlstyle\endcsname\relax
  \providecommand{\doi}[1]{doi: #1}\else
  \providecommand{\doi}{doi: \begingroup \urlstyle{rm}\Url}\fi

\bibitem[Bhat et~al.(2019)Bhat, Danelljan, et~al.]{bhat2019learning}
Goutam Bhat, Martin Danelljan, et~al.
\newblock Learning discriminative model prediction for tracking.
\newblock In \emph{Proc.\ IEEE Int.\ Conf.\ Computer Vision}, pages 6182--6191,
  2019.

\bibitem[Blatter et~al.(2023)Blatter, Kanakis, et~al.]{blatter2023efficient}
Philippe Blatter, Menelaos Kanakis, et~al.
\newblock Efficient visual tracking with exemplar transformers.
\newblock In \emph{IEEE Winter Conf. App. Computer Vision}, pages 1571--1581,
  2023.

\bibitem[Borsuk et~al.(2022)Borsuk, Vei, et~al.]{borsuk2022fear}
Vasyl Borsuk, Roman Vei, et~al.
\newblock {FEAR}: Fast, efficient, accurate and robust visual tracker.
\newblock In \emph{Proc.\ European Conf.\ Computer Vision}, pages 644--663.
  Springer, 2022.

\bibitem[Cao et~al.(2021)Cao, Fu, et~al.]{cao2021hift}
Ziang Cao, Changhong Fu, et~al.
\newblock Hi{FT}: Hierarchical feature transformer for aerial tracking.
\newblock In \emph{Proc.\ IEEE Int.\ Conf.\ Computer Vision}, pages
  15457--15466, 2021.

\bibitem[Chen et~al.(2021)Chen, Yan, et~al.]{chen2021transformer}
Xin Chen, Bin Yan, et~al.
\newblock Transformer tracking.
\newblock In \emph{Proc.\ IEEE Conf.\ Computer Vision Pattern Recognition},
  pages 8126--8135, 2021.

\bibitem[Chen et~al.(2019)Chen, Yang, et~al.]{chen2019detnas}
Yukang Chen, Tong Yang, et~al.
\newblock Detnas: Backbone search for object detection.
\newblock \emph{Advances in Neural Information Processing Systems}, 32, 2019.

\bibitem[Cui et~al.(2022)Cui, Jiang, et~al.]{cui2022mixformer}
Yutao Cui, Cheng Jiang, et~al.
\newblock Mixformer: End-to-end tracking with iterative mixed attention.
\newblock In \emph{Proc.\ IEEE Conf.\ Computer Vision Pattern Recognition},
  pages 13608--13618, 2022.

\bibitem[Dai et~al.(2022)Dai, Fu, et~al.]{dai2022siamese}
Jiahai Dai, Yunhao Fu, et~al.
\newblock Siamese hierarchical feature fusion transformer for efficient
  tracking.
\newblock \emph{Frontiers in Neurorobotics}, 2022.

\bibitem[Danelljan et~al.(2017)Danelljan, Bhat, et~al.]{danelljan2017eco}
Martin Danelljan, Goutam Bhat, et~al.
\newblock {ECO}: Efficient convolution operators for tracking.
\newblock In \emph{Proc.\ IEEE Conf.\ Computer Vision Pattern Recognition},
  pages 6638--6646, 2017.

\bibitem[Ding et~al.(2021)Ding, Zhang, et~al.]{ding2021repvgg}
Xiaohan Ding, Xiangyu Zhang, et~al.
\newblock Repvgg: Making vgg-style convnets great again.
\newblock In \emph{Proc.\ IEEE Conf.\ Computer Vision Pattern Recognition},
  pages 13733--13742, 2021.

\bibitem[Dosovitskiy et~al.(2021)Dosovitskiy, Beyer, et~al.]{dosovitskiy2021an}
Alexey Dosovitskiy, Lucas Beyer, et~al.
\newblock An image is worth 16x16 words: Transformers for image recognition at
  scale.
\newblock In \emph{International Conference on Learning Representations}, 2021.
\newblock URL \url{https://openreview.net/forum?id=YicbFdNTTy}.

\bibitem[Fan et~al.(2021)Fan, Bai, et~al.]{fan2021lasot}
Heng Fan, Hexin Bai, et~al.
\newblock La{SOT}: A high-quality large-scale single object tracking benchmark.
\newblock \emph{Int.\ J.\ Computer Vision}, 129:\penalty0 439--461, 2021.

\bibitem[He et~al.(2016)He, Zhang, et~al.]{he2016deep}
Kaiming He, Xiangyu Zhang, et~al.
\newblock Deep residual learning for image recognition.
\newblock In \emph{Proc.\ IEEE Conf.\ Computer Vision Pattern Recognition},
  pages 770--778, 2016.

\bibitem[Henriques et~al.(2015)Henriques, Caseiro, et~al.]{henriques2014high}
João~F. Henriques, Rui Caseiro, et~al.
\newblock High-speed tracking with kernelized correlation filters.
\newblock \emph{IEEE Trans.\ Pattern Anal.\ Machine Intell.}, 37\penalty0
  (3):\penalty0 583--596, 2015.
\newblock \doi{10.1109/TPAMI.2014.2345390}.

\bibitem[Huang et~al.(2021)Huang, Zhao, and Huang]{Huang2021}
Lianghua Huang, Xin Zhao, and Kaiqi Huang.
\newblock G{OT}-10k: A large high-diversity benchmark for generic object
  tracking in the wild.
\newblock \emph{IEEE Trans.\ Pattern Anal.\ Machine Intell.}, 43\penalty0
  (5):\penalty0 1562--1577, 2021.
\newblock \doi{10.1109/TPAMI.2019.2957464}.

\bibitem[Ioffe and Szegedy(2015)]{ioffe2015batch}
Sergey Ioffe and Christian Szegedy.
\newblock Batch normalization: Accelerating deep network training by reducing
  internal covariate shift.
\newblock In \emph{Proceedings of the 32nd International Conference on Machine
  Learning}, volume~37 of \emph{Proceedings of Machine Learning Research},
  pages 448--456, Lille, France, 07--09 Jul 2015. PMLR.

\bibitem[Javed et~al.(2022)Javed, Danelljan, et~al.]{9913708}
Sajid Javed, Martin Danelljan, et~al.
\newblock Visual object tracking with discriminative filters and siamese
  networks: A survey and outlook.
\newblock \emph{IEEE Trans.\ Pattern Anal.\ Machine Intell.}, pages 1--20,
  2022.
\newblock \doi{10.1109/TPAMI.2022.3212594}.

\bibitem[Kiani~Galoogahi et~al.(2017)Kiani~Galoogahi, Fagg,
  et~al.]{kiani2017need}
Hamed Kiani~Galoogahi, Ashton Fagg, et~al.
\newblock Need for speed: A benchmark for higher frame rate object tracking.
\newblock In \emph{Proc.\ IEEE Int.\ Conf.\ Computer Vision}, pages 1125--1134,
  2017.

\bibitem[Lin et~al.(2022)Lin, Fan, et~al.]{lin2022swintrack}
Liting Lin, Heng Fan, et~al.
\newblock Swintrack: A simple and strong baseline for transformer tracking.
\newblock \emph{Advances in {N}eural {I}nformation {P}rocessing {S}ystems},
  35:\penalty0 16743--16754, 2022.

\bibitem[Loshchilov and Hutter(2017)]{loshchilov2017sgdr}
Ilya Loshchilov and Frank Hutter.
\newblock {SGDR}: Stochastic gradient descent with warm restarts.
\newblock In \emph{International Conference on Learning Representations}, 2017.
\newblock URL \url{https://openreview.net/forum?id=Skq89Scxx}.

\bibitem[Loshchilov and Hutter(2019)]{loshchilov2018decoupled}
Ilya Loshchilov and Frank Hutter.
\newblock Decoupled weight decay regularization.
\newblock In \emph{International Conference on Learning Representations}, 2019.
\newblock URL \url{https://openreview.net/forum?id=Bkg6RiCqY7}.

\bibitem[Mehta and Rastegari(2022)]{mehta2022mobilevit}
Sachin Mehta and Mohammad Rastegari.
\newblock Mobilevit: Light-weight, general-purpose, and mobile-friendly vision
  transformer.
\newblock In \emph{International Conference on Learning Representations}, 2022.
\newblock URL \url{https://openreview.net/forum?id=vh-0sUt8HlG}.

\bibitem[Muller et~al.(2018)Muller, Bibi, et~al.]{muller2018trackingnet}
Matthias Muller, Adel Bibi, et~al.
\newblock Trackingnet: A large-scale dataset and benchmark for object tracking
  in the wild.
\newblock In \emph{Proc.\ European Conf.\ Computer Vision}, pages 300--317,
  2018.

\bibitem[Paszke et~al.(2019)Paszke, Gross, et~al.]{paszke2019pytorch}
Adam Paszke, Sam Gross, et~al.
\newblock Pytorch: An imperative style, high-performance deep learning library.
\newblock \emph{Advances in Neural Information Processing Systems}, 32, 2019.

\bibitem[Sandler et~al.(2018)Sandler, Howard, et~al.]{sandler2018mobilenetv2}
Mark Sandler, Andrew Howard, et~al.
\newblock Mobilenetv2: Inverted residuals and linear bottlenecks.
\newblock In \emph{Proc.\ IEEE Conf.\ Computer Vision Pattern Recognition},
  pages 4510--4520, 2018.

\bibitem[Tay et~al.(2022)Tay, Dehghani, et~al.]{tay2022efficient}
Yi~Tay, Mostafa Dehghani, et~al.
\newblock Efficient transformers: A survey.
\newblock \emph{ACM Computing Surveys}, 55\penalty0 (6):\penalty0 1--28, 2022.

\bibitem[Vaswani et~al.(2017)Vaswani, Shazeer, et~al.]{vaswani2017attention}
Ashish Vaswani, Noam Shazeer, et~al.
\newblock Attention is all you need.
\newblock \emph{Advances in Neural Information Processing Systems}, 30, 2017.

\bibitem[Wang et~al.(2015)Wang, Shi, et~al.]{wang2015understanding}
Naiyan Wang, Jianping Shi, et~al.
\newblock Understanding and diagnosing visual tracking systems.
\newblock In \emph{Proc.\ IEEE Int.\ Conf.\ Computer Vision}, pages 3101--3109.
  IEEE, 2015.

\bibitem[Wu et~al.(2021)Wu, Xiao, et~al.]{wu2021cvt}
Haiping Wu, Bin Xiao, et~al.
\newblock Cv{T}: Introducing convolutions to vision transformers.
\newblock In \emph{Proc.\ IEEE Int.\ Conf.\ Computer Vision}, pages 22--31,
  2021.

\bibitem[Yan et~al.(2021{\natexlab{a}})Yan, Peng, et~al.]{yan2021learning}
Bin Yan, Houwen Peng, et~al.
\newblock Learning spatio-temporal transformer for visual tracking.
\newblock In \emph{Proc.\ IEEE Int.\ Conf.\ Computer Vision}, pages
  10448--10457, 2021{\natexlab{a}}.

\bibitem[Yan et~al.(2021{\natexlab{b}})Yan, Peng, et~al.]{yan2021lighttrack}
Bin Yan, Houwen Peng, et~al.
\newblock Light{T}rack: Finding lightweight neural networks for object tracking
  via one-shot architecture search.
\newblock In \emph{Proc.\ IEEE Conf.\ Computer Vision Pattern Recognition},
  pages 15180--15189, 2021{\natexlab{b}}.

\bibitem[Yan et~al.(2021{\natexlab{c}})Yan, Zhang, et~al.]{yan2021alpha}
Bin Yan, Xinyu Zhang, et~al.
\newblock Alpha-refine: Boosting tracking performance by precise bounding box
  estimation.
\newblock In \emph{Proc.\ IEEE Conf.\ Computer Vision Pattern Recognition},
  pages 5289--5298, 2021{\natexlab{c}}.

\bibitem[Ye et~al.(2022)Ye, Chang, et~al.]{ye2022joint}
Botao Ye, Hong Chang, et~al.
\newblock Joint feature learning and relation modeling for tracking: A
  one-stream framework.
\newblock In \emph{Proc.\ European Conf.\ Computer Vision}, pages 341--357.
  Springer, 2022.

\bibitem[Čehovin et~al.(2016)Čehovin, Leonardis, and
  Kristan]{vcehovin2016visual}
Luka Čehovin, Aleš Leonardis, and Matej Kristan.
\newblock Visual object tracking performance measures revisited.
\newblock \emph{IEEE Trans.\ Image Process.}, 25\penalty0 (3):\penalty0
  1261--1274, 2016.
\newblock \doi{10.1109/TIP.2016.2520370}.

\end{thebibliography}
	
\end{document}